\documentclass[10pt,twocolumn,letterpaper]{article}

\usepackage{iccv}
\usepackage{times}
\usepackage{epsfig}
\usepackage{graphicx}
\usepackage{amsmath}
\usepackage{amssymb}

\usepackage[breaklinks=true,bookmarks=false]{hyperref}

\iccvfinalcopy 

\def\iccvPaperID{****} 

\ificcvfinal\fi
\begin{document}

\title{Hand3D: Hand Pose Estimation using 3D Neural Network}

\author{Xiaoming Deng$^{1*}$~~Shuo Yang$^{1*}$~~Yinda Zhang$^{2}$\thanks{indicates equal contribution.} ~~ Ping Tan$^3$~~Liang Chang$^4$~~Hongan Wang$^1$ \\ \\
$^1$Institute of Software, CAS~~~$^2$Princeton University~~~$^3$Simon Fraser University~~~$^4$Beijing Normal University\\
}

\maketitle

\begin{abstract}
  We propose a novel 3D neural network architecture for 3D hand pose estimation from a single depth image. Different from previous works that mostly run on 2D depth image domain and require intermediate or post process to bring in the supervision from 3D space, we convert the depth map to a 3D volumetric representation, and feed it into a 3D convolutional neural network(CNN) to directly produce the pose in 3D requiring no further process. Our system does not require the ground truth reference point for initialization, and our network architecture naturally integrates both local feature and global context in 3D space. To increase the coverage of the hand pose space of the training data, we render synthetic depth image by transferring hand pose from existing real image datasets. We evaluation our algorithm on two public benchmarks and achieve the state-of-the-art performance. The synthetic hand pose dataset will be available.
\end{abstract}

\section{Introduction}

Hand pose estimation is ubiquitously required in many critical applications to support human computer interaction, such as autonomous driving, virtual/mixed reality, and robotics \cite{Erolsurvey}. Given a single image of a hand, we would like to estimate the 3D hand pose, i.e. the location of each hand joint in 3D space. While many early works \cite{athitsos2003,wu2005analyzing,wang2009real} took color images as input, methods built upon depth images usually exhibit superior performance \cite{oikonomidis2011full}, because depth provides essential 3D information that is extreme helpful to localize joints. In this paper, we focus on estimating 3D hand pose from a single depth image.

This task is known to be extremely challenging because of the severe occlusions caused by articulate hand pose, and the noisy input from affordable depth sensors. 
Most of the previous methods \cite{tompson14tog,ge2016robust} estimates the location of each hand joint on 2D depth image domain, followed by a post-processing in which hand joints are projected to 3D space and optimized by a predefined hand model to reduce error that is significant in 3D space but not apparent in 2D image domain. Such post-process has been thoroughly studied but far from optimal, and \cite{supancic2015depth} shows that even a directly searching of nearest neighbors pose in 3D space could also achieve performance close to the state of the art, which shows the benefits of directly solving pose estimation in 3D space.
\begin{figure*}[t]
\centering
\includegraphics[width=\textwidth]{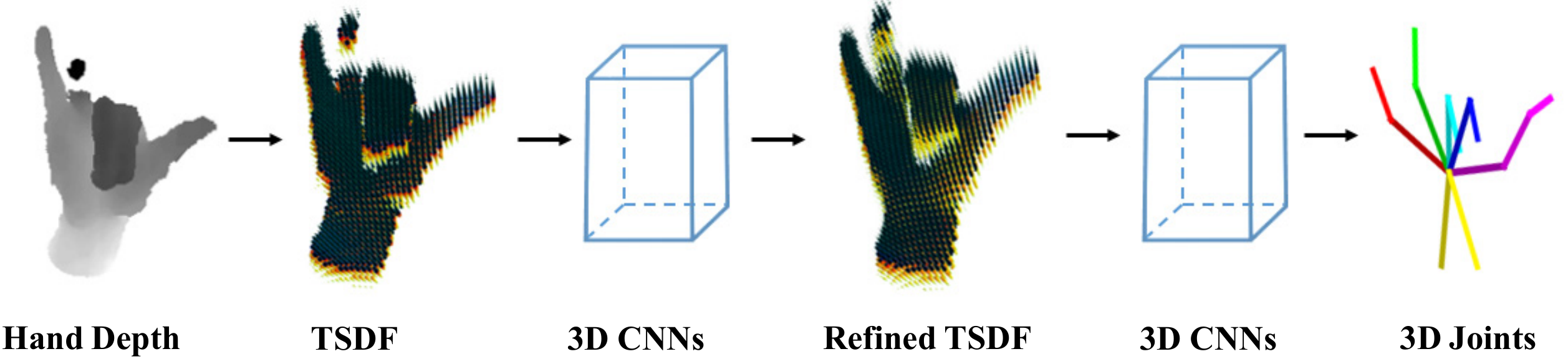}
\caption{The whole system pipeline. We firstly compute the TSDF volumetric feature with a depth image, use a 3D convolutional neural network to get the refined TSDF, finally feed the refined TSDF feature to recover the 3D positions of hand joints.}
\label{fig:fig1}
\end{figure*}

Inspired by \cite{supancic2015depth}, we propose a 3D convolutional neural network architecture for 3D hand pose estimation (Refer to Fig.~\ref{fig:fig1}).
The input depth images are firstly aligned with respect to the center of mass (COM), so that we do not require perfect hand alignment as initial.
The aligned input depth image is then converted to a 3D volumetric representation, and a 3D convolutional neural network(CNN) is trained directly on it to estimate the location of each hand joint in the COM coordinate.
Our network naturally learns to use both the 3D feature and the 3D global context for hand pose estimation. Hence, the output of our network satisfies the context prior and does not require any further post-processing to integrate context in predefined hand model.

The unsatisfactory quality and quantity of the data is yet another problem that makes models learned for hand pose estimation unreliable. The depth image from affordable/portable sensor usually suffers from noise and missing depth. On the other hand, the relatively small training set collected from a few subjects could hardly cover either the pose space or different hand configurations, e.g. bone length. To solve these issues, we perform data augmentation to improve both the quality and quantity of the data during both training and testing. We train a fully convolutional network (FCN) to complete the hand shape from a single depth view. Empirically we find this network significantly improves the quality of the input depth, e.g. filling up holes, synthesizing occluded parts. To have a large training data with better coverage of the hand configuration, we transfer hand pose from existing dataset \cite{tompson14tog} to handle with variable bone lengths, and render the depth accordingly. Both these two approaches of data augmentation brings significant improvement to pose accuracy. We evaluate our method on two popular datasets, and demonstrate state-of-the-art performance on hand pose estimation.

The contributions of our method are mainly two aspects. 1) We propose a 3D network to directly estimate the 3D hand pose. Our method does not rely on any predefined model and require no post-processing for 2D/3D projection which may potentially increase error. 2) We perform data augmentation to increase both the quality and quantity of the training data, which are essential to achieve good performance. We use a 3D FCN to refine the TSDF which completes missing depth. We also learn poses from real datasets and transfer them to synthetic CAD models.

\section{Related Work}
Hand pose estimation has been extensively studied in many previous works, and comprehensive review on color image and depth image based hand pose estimation are given in Erol \etal \cite{Erolsurvey} and Supancic \etal \cite{supancic2015depth}.
With a plenty amount of training data, hand pose can be directly learned, e.g. using random forest \cite{tang2014latent}, or retrieved, e.g. KNN \cite{athitsos2003}.
To handle the heavy occlusion, 3D hand model has been used to bring context regulation and refine the result \cite{wang2013video,oikonomidis2011full,wang2009real}.
However, none of these works effectively took advantage of the large scale training data with the state of the art learning technique.

Recently, convolutional neural network has been demonstrated to be effective in handle articulated pose estimation.
The hand skeleton joint locations are estimated in the depth image domain as a heat map via classification \cite{tompson14tog,ge2016robust}, or directly by regression \cite{oberweger15, zhou2016model}.
However, to produce final result in 3D space, the intermediate result learned on 2D image domain has to be projected to 3D during the learning procedure or post process.
DeepPrior\cite{oberweger15} used CNN to regress the hand skeleton joints by exploiting pose priors. DeepModel\cite{zhou2016model} proposed a deep learning approach with a new forward kinematics based layer, which helps to ensure the geometric validity of estimated poses.  Oberweger et. al. \cite{Oberweger15a} used a feedback loop for hand pose estimation and depth map synthesizing to refine hand pose iteratively.  However, these regression models either require a careful initial alignment or being sensitive to the predefined bone length. In contrast, we directly convert the input to 3D volumetric and perform all computation in 3D to prevent potential error.

3D deep learning has been used for 3D object detection \cite{DeepSlidingShapes} and scene understanding \cite{zhang2016deepcontext}.
We extend the idea for 3D hand pose estimation.
The most related work to our approach is \cite{supancic2015depth}, which proposed to directly estimate hand pose in 3D space using nearest neighbor search.
We apply deep learning technique to better leverage 3D evidence.

Deep learning model for hand pose estimation needs to see training data of hand images with a variety of poses, bone length.
However, available datasets often contain biased hand pose from a few number of subjects \cite{tompson14tog}.
Therefore, data augmentation is important.
Though expensive, commercial digital gloves have been used to get hand motion data, and the rigged hand mesh model is rendered as training images \cite{wang2009real,bellon2016model}.
Comparatively affordable, model based tracking  \cite{tompson14tog} and 2D based annotation \cite{Oberweger16} have been used to generate synthetic data more efficiently.
However, the quality of the label highly depends on the algorithm, which may not be reliable sometimes.
We propose to directly transfer the hand pose from other annotated dataset to a hand CAD model with changeable bone length.

\section{Approach}
\subsection{Overview}

The overall pipeline of our system is illustrated in Fig.~\ref{fig:fig1}.
Given a depth image of a hand as input, we first build a reference coordinate at the center of mass (COM) of the foreground region.
We convert the input depth to a truncated signed distance function (TSDF \cite{kinectfusion}) representation built on this coordinate and feed it into the TSDF refinement network, which removes the artifacts caused by noisy and missing depth.
The refined TSDF is then fed into the 3D pose network to estimate the 3D location of each hand skeleton joint relative to the COM.
As the input and computation are all in 3D, our system learns 3D context for pose estimation, and therefore does not require any post-processing or external model for regulation, and thus runs efficiently.

\subsection{3D Pose Estimation Network}
\label{sec:parameter}
\begin{figure}[t]
\centering
\begin{tabular}{cc}
\includegraphics[width=0.3\linewidth]{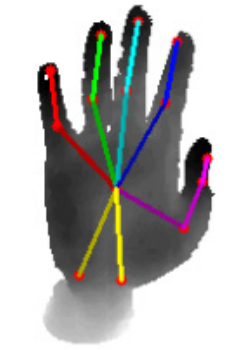} & \includegraphics[width=0.35\linewidth]{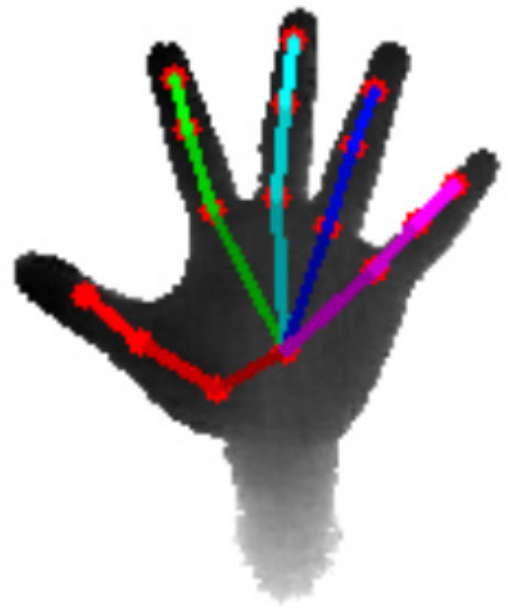} \\
(a) NYU hand model & (b) ICVL hand model \\
\end{tabular}
\caption{Illustration of our hand skeleton model. (a) shows hand joints in NYU hand pose dataset\cite{tompson14tog}, which contains 14 joints. (b) shows hand joints in ICVL hand pose dataset\cite{tang2014latent}, which contains 16 joints.}
\label{fig:fig3}
\end{figure}

\noindent \textbf{Hand pose parametrization.}
We use the location of each hand skeleton joint to represent the hand pose, as shown in Fig. \ref{fig:fig3}.
For visualization purpose, the lines connecting two neighboring joints, representing bones, are also shown but not used in our model.
we experiments on two widely used hand pose benchmarks NYU hand dataset \cite{tompson14tog} and ICVL hand dataset \cite{tang2014latent}. Although they adopt the hand model with different numbers of joints and connection topology, our method can be easily applied since we do not rely on any pre-defined structural hand model.

\begin{figure}[h]
\centering
\includegraphics[width=0.48\textwidth]{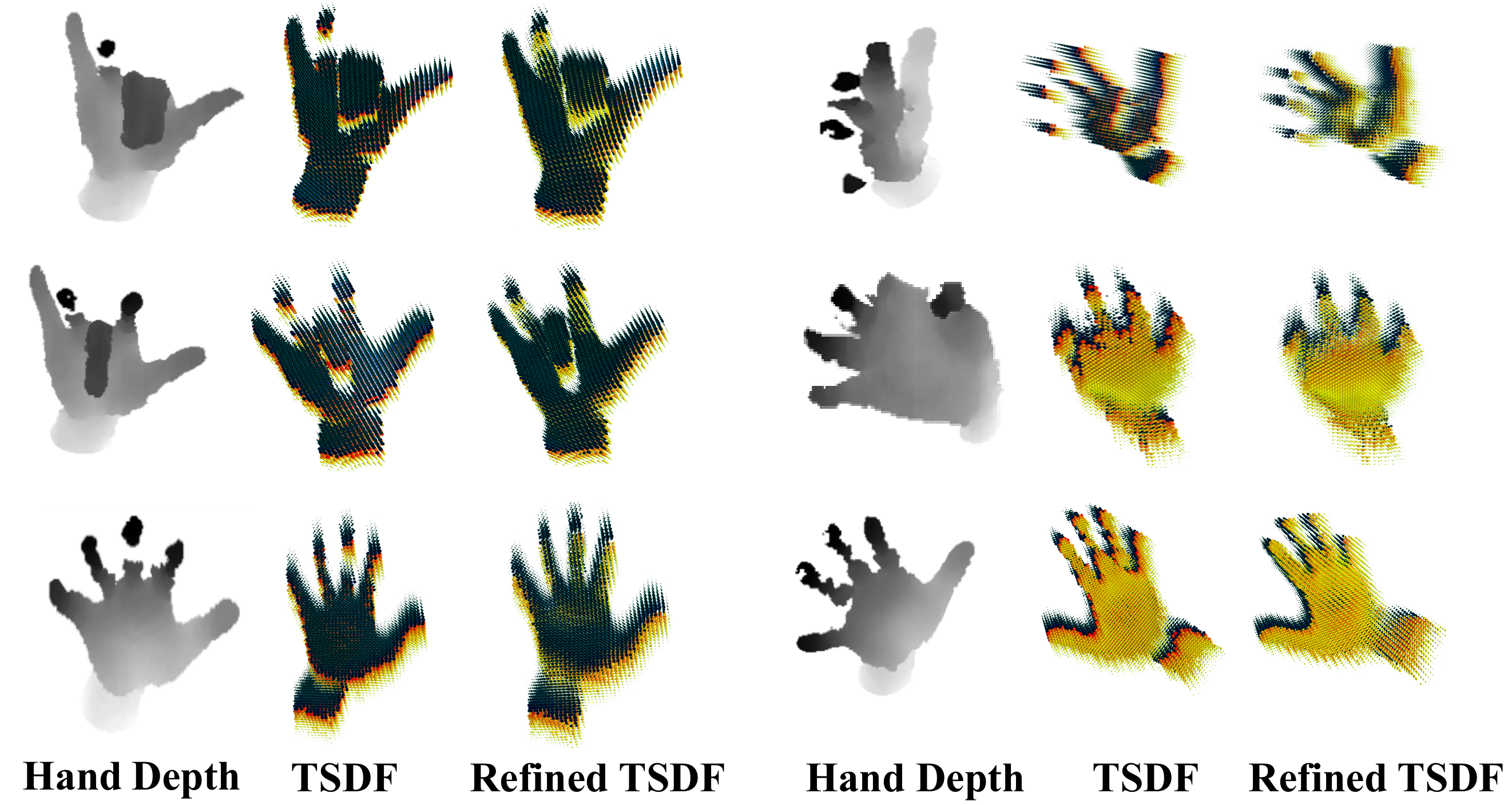}
\caption{Illustration of 3D volumetric representation TSDF and refinement. Hand depth map is converted into a volumetric feature by using a truncated signed distance function(TSDF), and the feature is refined with TSDF refinement network. We can observe that the refined TSDF has less missing data and artifact than the original TSDF.}
\label{fig:fig3b}
\end{figure}

\noindent \textbf{3D volumetric representation.} We convert the input depth image into a truncated signed distance function(TSDF) representation \cite{kinectfusion}, which has advantages over depth map and mesh in encoding uncertainty in the range data, handling missing depth, and implicitly embedding surface geometry.
We first find the foreground region by hand segmentation, and then calculate the center of mass (COM).
We align this COM to the origin of a reference coordinate, and calculate a TSDF with the resolution of $60\times60\times60$ voxels.
Each voxel represents a space of $5\times5\times5$ mm, and the whole TSDF expands a space of $300\times300\times300$ mm, which is sufficient for most of the poses.
The truncation value of TSDF is set to 50mm.  Fig. \ref{fig:fig3b} shows some examples of input depths and its corresponding TSDF.

\begin{figure}[t]
\centering
\includegraphics[width=0.5\textwidth]{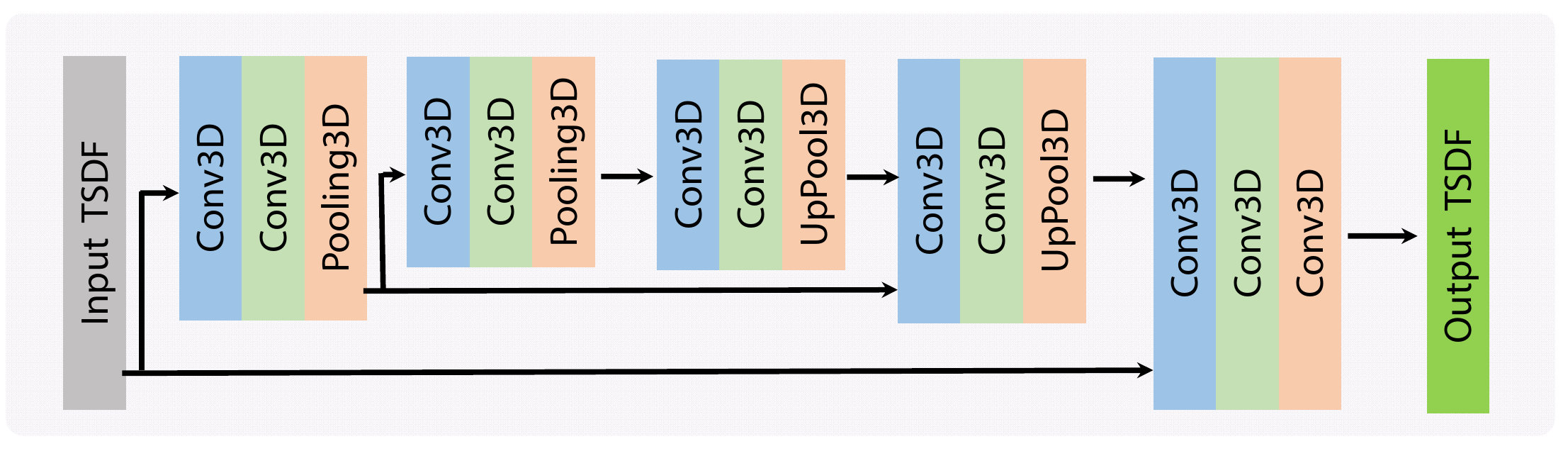}
\caption{Network architecture for TSDF refinement. This network is a 3D fully convolutional network. Taking the TSDF of a raw depth as input. the network produces a refined TSDF, which will be used as the input to the pose estimation network. }
\label{fig:figtsdf}
\end{figure}

\noindent \textbf{Hand shape refinement.} 
The quality of the depth combined from multiple depth sensors is significantly higher than that of a single raw depth from affordable sensor like kinect.
NYU dataset provides well aligned pairs of combined depth and raw depth.
Taking the TSDF of the raw depth as input, we train a 3D fully convolutional network (FCN) to estimate the TSDF of the combined depth. The architecture of our 3D FCN is shown in Fig. \ref{fig:figtsdf}. The network consists of an encoder with two layers of 3D convolution + 3D convolution + 3D pooling (with ReLU), and a symmetric decoder with two layers of 3D convolution + 3D convolution + 3D uppooling, followed by three 3D convolution. We also add the short cut link in the network to bring the high-dimensional feature map from the encoder to the decoder to maintain high resolution details. Empirically, we find this TSDF refinement consistently helps to improve the performance of pose estimation.
Fig.~\ref{fig:fig3b} shows some qualitative results of our shape refine network.
We can see that the network is effective in completing the missing depth and removing artifacts caused by noisy raw depths.


\begin{figure}[t]
\centering
\includegraphics[width=0.5\textwidth]{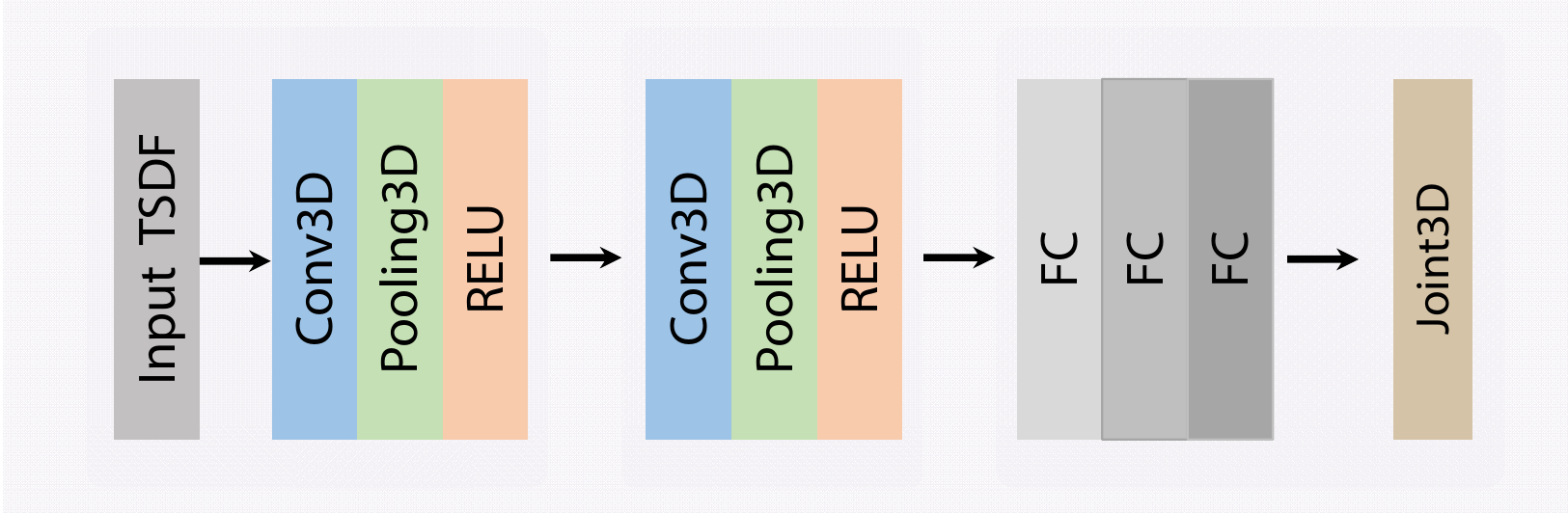}
\caption{Network architecture for hand pose estimation. The network directly estimates the 3D location of each joint relative to the center of mass taking the refined TSDF as input. }
\label{fig:fig2}
\end{figure}

\noindent \textbf{Pose estimation network.}
Our 3D ConvNet architecture for pose estimation is shown in Fig. \ref{fig:fig2}.
The network starts with two layers of 3D convolution + 3D pooling + ReLU, followed by three fully connected layers. The last fully connected layer produces a feature of dimension $3\times$ \# of joints, which is used to directly estimate the 3D location of each joint relative to the COM. We use a L2-norm loss layer, which computes the euclidean distance between the prediction of 3D hand joint positions and the ground truth.

\subsection{Data Augmentation}
Hand pose datasets are usually collected with a small number of subjects, like two users for NYU hand pose dataset \cite{tompson14tog}, such that fails to include hands with different configurations, e.g. bone length, skinning. Some poses are also unintentionally overlooked due to the habits of the data collectors. To overcome these limitations, we extract hand pose from NYU dataset, and transfer them to a configurable hand CAD model. We render depth images for hands with different bone lengths from multiple viewpoints. Compared to previous data augmentation work for hand pose estimation where the pose are created heuristically or captured by error-prone data glove \cite{xu2013efficient,wang2009real,bellon2016model}, our pose are extracted from real data.

\noindent \textbf{Pose parametrization for synthetic data.}
Different from the pose representation introduced in Section~\ref{sec:parameter}, we use the angle between two connecting bones on each joint to represent the hand. Changing the bone length while maintaining these angles is likely to produce natural looking hand poses.
The hand pose is parametrized similar to \cite{wang2013video,wang2009real} with position and orientation of root joint, relative joint angles of each joints. In NYU hand skeleton model, we use palm center joint as the root joint, and use 30 degrees of freedom (DOF): 6 DOF for root pose, 4 DOF for thumb, and 5 DOF for each of other finger. 

\noindent \textbf{Learning pose from real dataset.}
Since the NYU and ICVL dataset only provide ground truth for hand skeleton joint locations, we need to recover the hand pose using inverse kinematics \cite{parent2012computer}.
Specifically, we optimize the parametrized hand pose $\mathbf{p}$ for the smallest forward kinematics cost, which measures the pairwise distances between the hand joint positions in hand model and the corresponding ground truth hand joints:
\begin{equation}
\label{eq1}
\mathbf{p} = \arg \min \frac{1}{n}\sum\limits_{i = 1}^n {\left\| {O_i - R_i (\mathbf{p})} \right\|}
\end{equation}
\noindent where $\{O_i\}_{i=1}^n$ are the coordinates of ground truth hand joint locations, $\{R_i\}_{i=1}^n$ are the coordinates of hypothesized hand joint locations, which are computed by hand skeleton model and forward kinematics with pose $\mathbf{p}$, $n$ is the number of hand joints. $\mathbf{p}$ consists in hand root joint position and relative angles of individual joints.

\begin{figure*}[t]
 \centering
 \begin{tabular}{cc}
\includegraphics[width=0.5\linewidth]{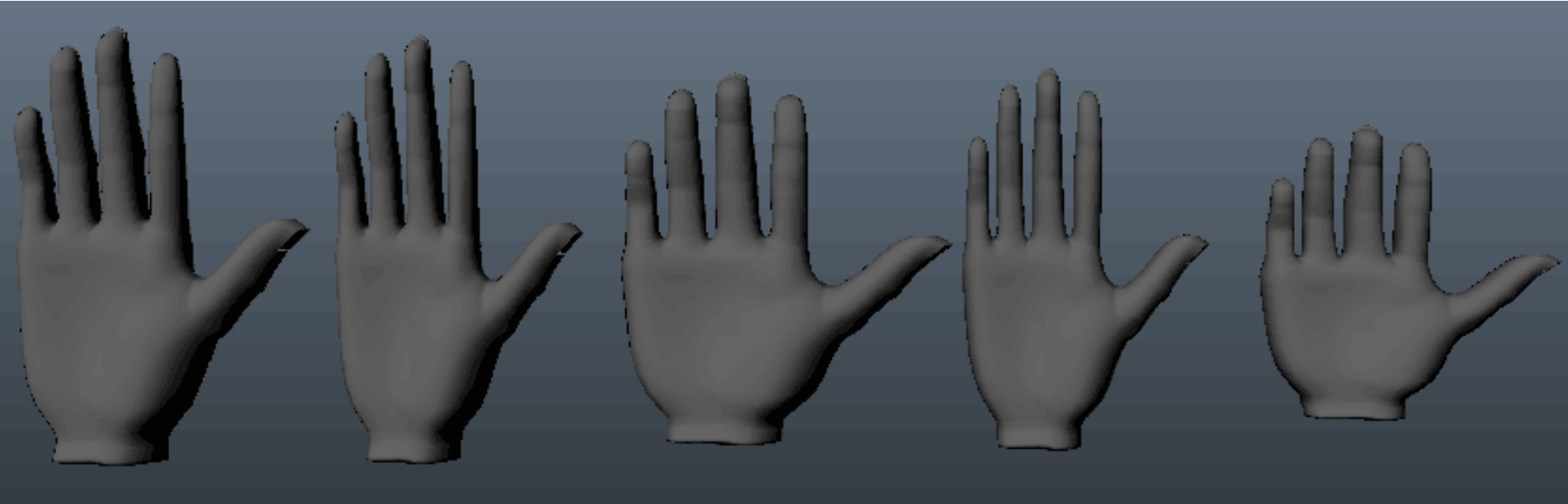} &
\includegraphics[width=0.35\linewidth]{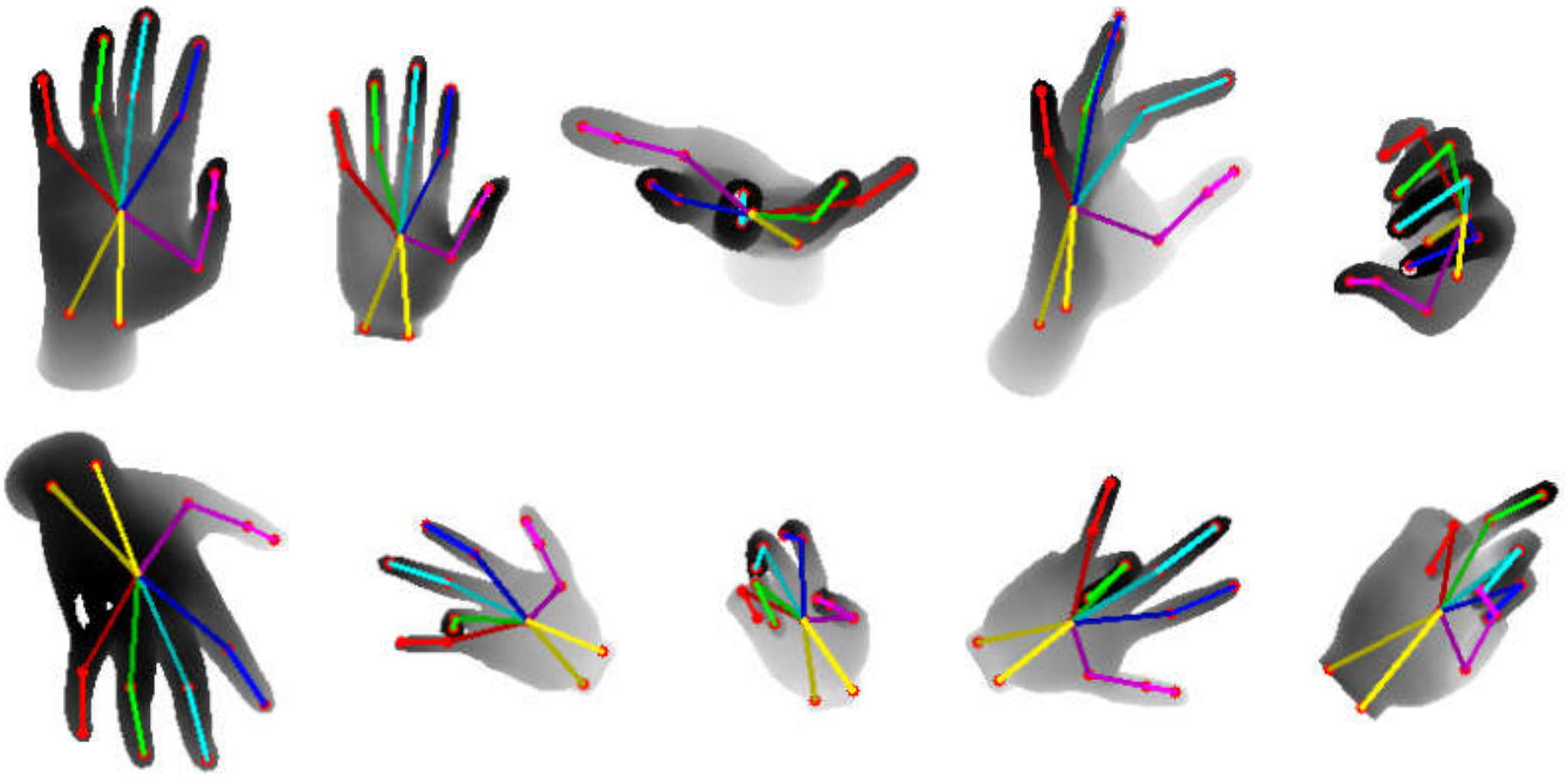}\\
(a) Base hand models  & (b) Synthetic depth image and ground truth skeleton joints
\end{tabular}
\caption{Data Augmentation. (a) shows base hand models with different hand skeleton sizes. (b) shows examples of the synthetic depth image and ground truth hand skeleton joints. The augmented data is generated with different sizes of human hand models. }
\label{fig:fig4}
\end{figure*}

\noindent \textbf{Transfer pose to synthetic data.}
We encode the recovered hand pose into a standard Biovision hierarchical data (BVH) file \cite{bvh}, which consists of the position of root joint, relative angles of each joint, and canonical bone lengths. We modify the relative offsets in the BVH file to generate models with different bone lengths, use linear skinning techniques \cite{baran2007automatic} to animate the chosen 3D hand CAD model given the pose, and then render depth images with commercial software Autodesk Maya. The ground truth position of hand joints can be computed by forward kinematics \cite{parent2012computer} and skeleton size. In building this data augmentation pipeline, we attempted to generate synthetic data with different skeleton sizes and different camera poses. We use 5 hand models as shown in Fig. \ref{fig:fig4}(a). Fig. \ref{fig:fig4}(b) illustrates several examples of synthetic depth images and ground truth hand joints. The camera pose is chosen randomly within a range to be representative of indoor hand gesture interaction scenario.

\section{Experiment}
\subsection{Dataset and Evaluation}
We evaluate our method on widely-used NYU hand dataset\cite{tompson14tog} and ICVL hand dataset \cite{tang2014latent} as comparatively larger variation of poses and hand shapes are exhibited in these two datasets. NYU dataset contains 72,757 training and 8,252 testing images, captured by PrimeSense camera. Ground truth joints are annotated using an accurate offline particle swarm optimization (PSO) algorithm, similar to \cite{oikonomidis2011full}.
The ICVL dataset \cite{tang2014latent} has over 300,000 training depth images and 2 testing sequences, with each about 800 frames. The depth images of ICVL dataset are captured by Intel Creative Interactive Gesture Camera.
Other popular hand datasets with relatively fixed poses (gesture), such as MSRA2015 \cite{sun2015cascaded}, donot satisfy our needs of evaluation for articulated pose estimation.

We evaluate the hand joint estimation performance using standard metrics proposed in \cite{tang2014latent}, which is widely used in many hand pose estimation work such as \cite{sun2015cascaded,oberweger15,supancic2015depth}. Specifically, we calculate the percentage of test examples that have all predicted joints within a given maximum distance from the ground truth. As stated in Supancic \etal \cite{supancic2015depth}, human annotators could achieve an accuracy about 20mm, and can easily distinguish two poses if the error is above 50mm. Therefore, we show the percentage of good frames under 20, 40, and 50mm error thresholds in our evaluation.

\subsection{Implementation Details}
Our model is implemented in the deep learning library Keras \cite{Keras}. The whole network (refine network + pose estimation network) is trained for 30 epochs with a mini-batch of 6 images, which takes about 5 GB RAM. The learning rate is set to 0.0001.

To have a fair comparison with DeepPrior \cite{oberweger15} and DeepModel \cite{zhou2016model}, we follow Oberweger \etal's approach \cite{oberweger15} to preprocess the raw depth, which is used for training and testing of all models. It is noted that we only assume the foreground mask of the hand area is known, but do not allow the use of the specific location of a reference point, e.g. the palm joint.
We believe such setting is more realistic for real application. For DeepPrior and DeepModel, the hand region during the testing stage is aligned with an estimated reference joint, which is obtained by regressing the offset from COM by using CNN, and get hand pose with the aligned depth image. The dimension of pose PCA embedding in DeepPrior is set to 30.

\subsection{Comparison with State-of-the-Art}
\begin{figure}[ht]
\centering
\begin{tabular}{cc}
\includegraphics[width=\linewidth]{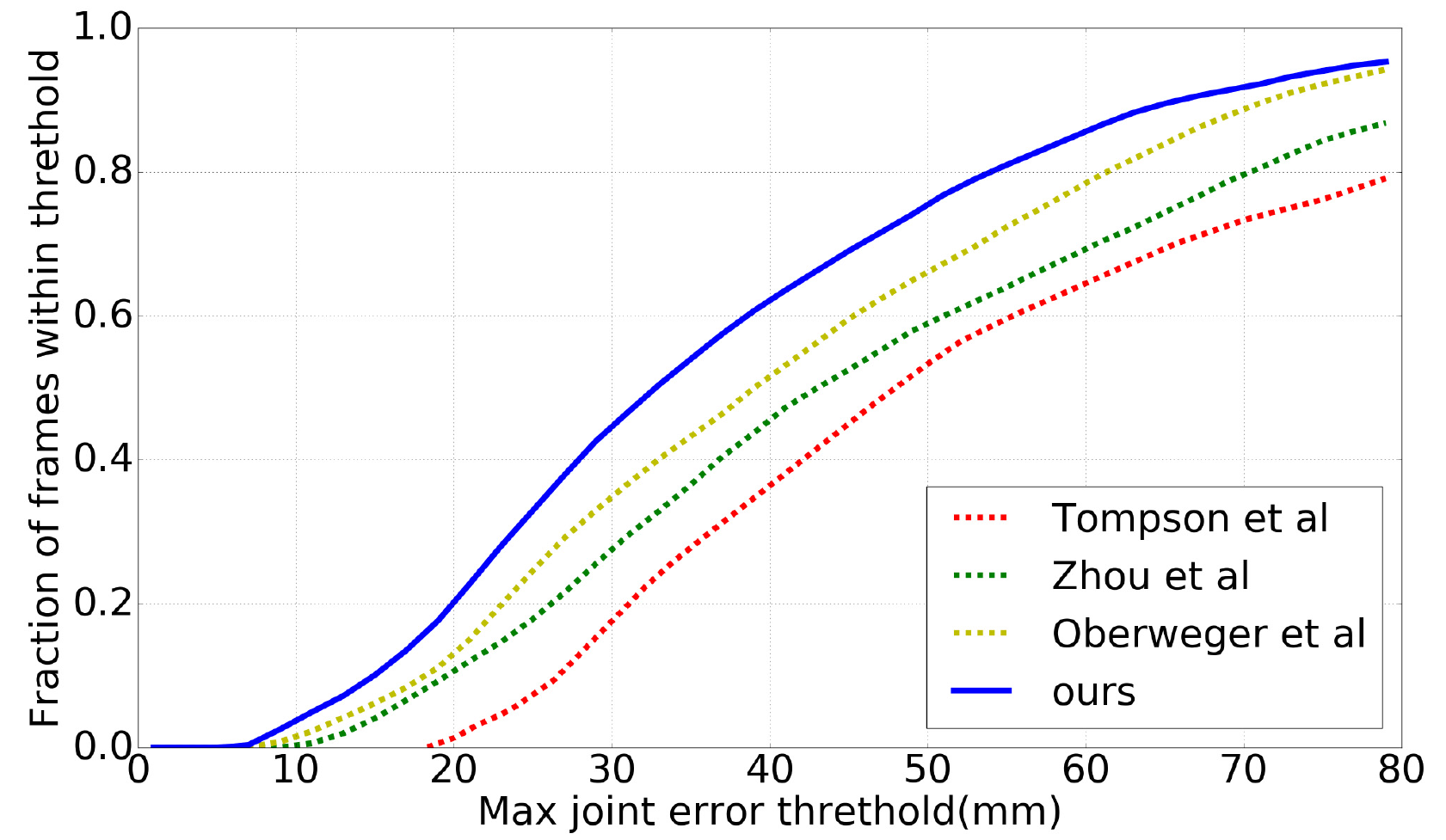} \\
(a) \\
\includegraphics[width=\linewidth]{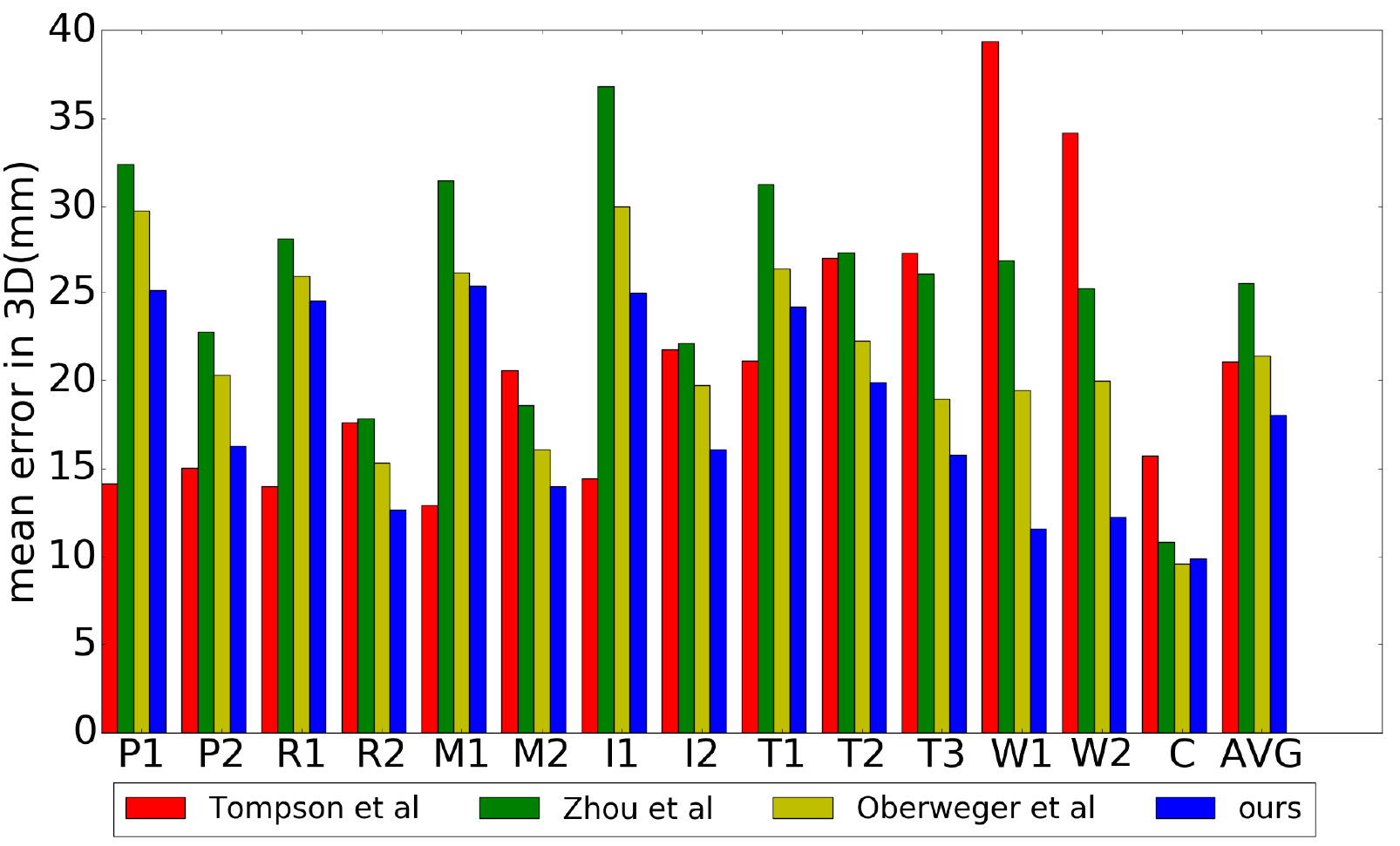}\\
(b)
\end{tabular}
\caption{Comparison with state-of-the-art methods on NYU hand pose dataset. (a) Percentage of frames in the test examples under different error thresholds. (b) Mean error distance for each joint across all the test examples. The palm and fingers are indexed as C: palm, T: thumb, I: index, M: middle, R: ring, P: pinky, W: wrist.}
\label{fig:fig6}
\end{figure}

\begin{figure}[ht]
\centering
\begin{tabular}{cc}
\includegraphics[width=\linewidth]{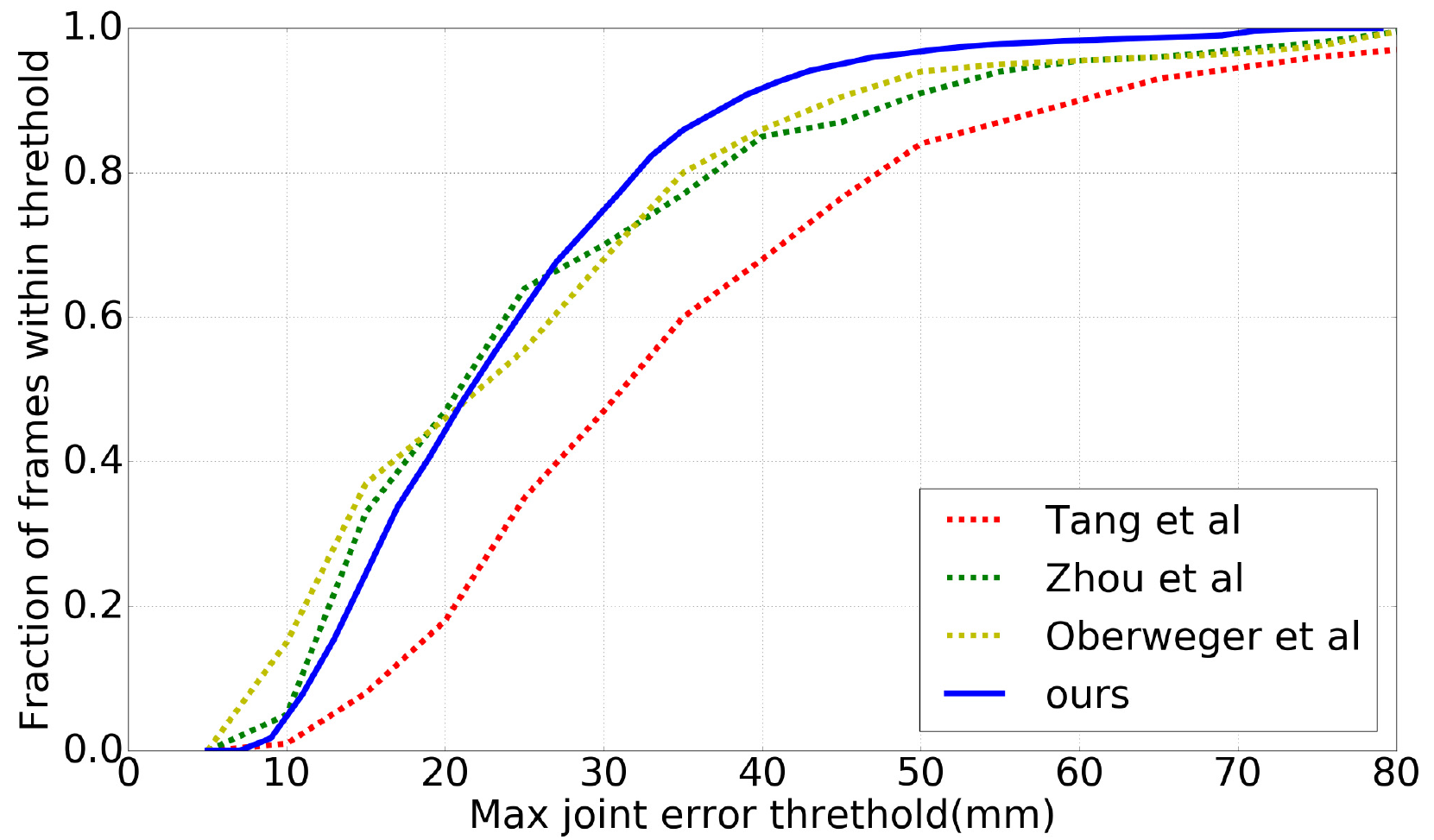} \\
(a) \\
\includegraphics[width=\linewidth]{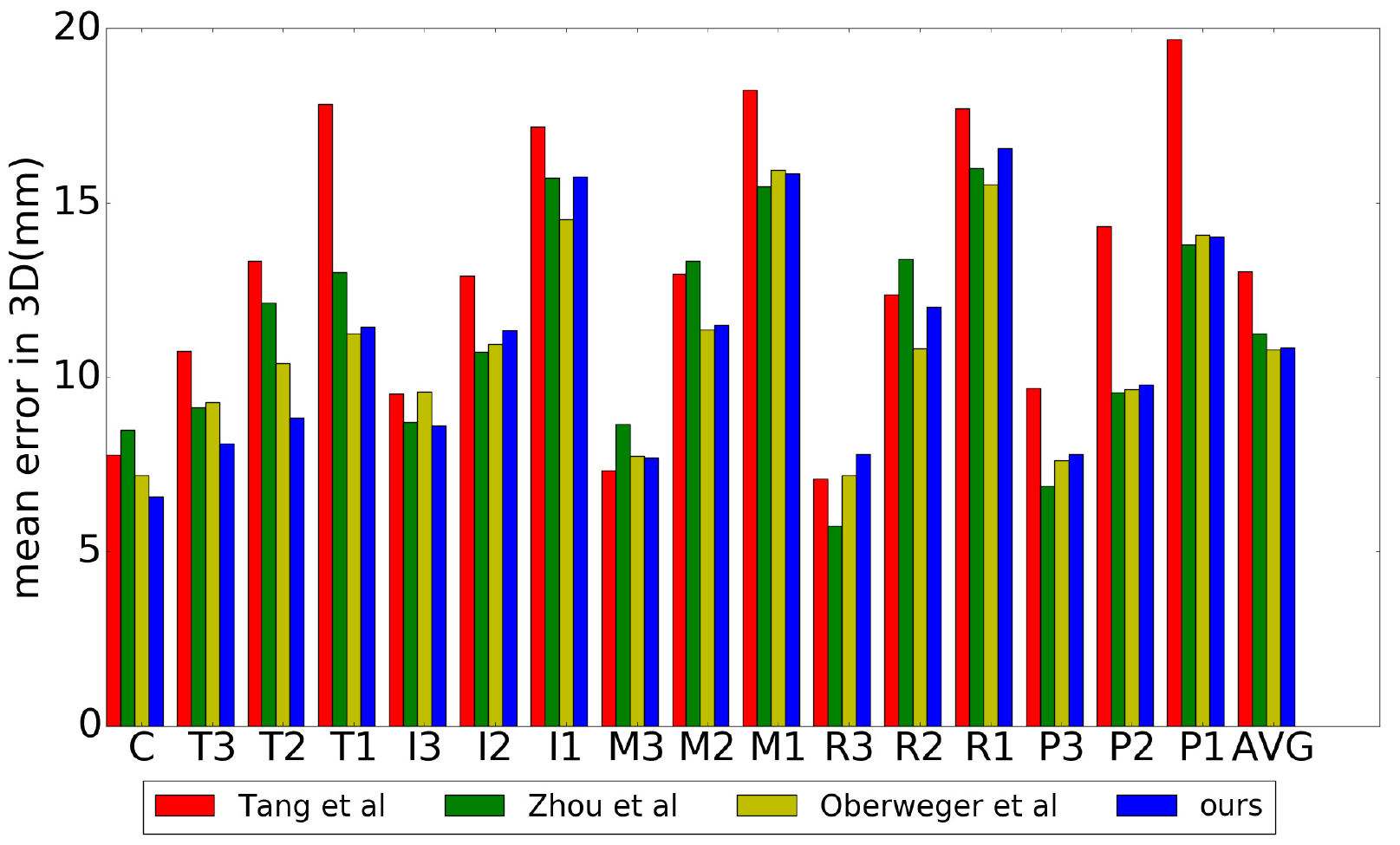}\\
(b)
\end{tabular}
\caption{Comparison with state-of-the-art methods on ICVL hand pose dataset. (a) Percentage of good test frames in the test examples under different error thresholds. (b) Mean error distance for each joint across all the test examples. The palm and fingers are indexed as C: palm, T: thumb, I: index, M: middle, R: ring, P: pinky, W: wrist.}
\label{fig:fig7}
\end{figure}

We compare our method to state-of-the-art methods on both NYU dataset and ICVL dataset.
One of the primary goals is to show that our method can be well adapted to different hand skeleton structures.
Quantitative results are shown in Fig. \ref{fig:fig6} and Fig. \ref{fig:fig7}.

On NYU dataset, we compare our method to three competitive methods: Tompson \etal \cite{tompson14tog}, Zhou \etal \cite{zhou2016model}, and Oberweger \etal \cite{oberweger15}, and results are shown in Fig. \ref{fig:fig6} and Table \ref{tab:comparenyu}.
In Fig. \ref{fig:fig6} (a), we show the percentage of testing examples (Y-axis) with the max joint error below thresholds (X-axis). We achieve consistent improvement over all error thresholds compared to other methods.
Fig. \ref{fig:fig6} (b) shows the mean error of each skeleton joint. As we can see the mean error distance for all joints with our method is 17.6mm, which is 4mm smaller than the results with Oberweger \etal \cite{oberweger15} and Tompson \etal \cite{tompson14tog}, and 8mm smaller than the results with Zhou \etal \cite{zhou2016model}. Table \ref{tab:comparenyu} shows statistics on several key error thresholds, and consistently our method achieves significantly better performance than all other methods.

On ICVL dataset, we compare our method to three state-of-the-art methods: Tang \etal \cite{tang2014latent}, Zhou \etal \cite{zhou2016model}, and Oberweger \etal \cite{oberweger15}, and the results are show in Fig. \ref{fig:fig7} and Table \ref{tab:compareicvl}. In Fig. \ref{fig:fig7} (a), we show the percentage of testing examples (Y-axis) with the max joint error below thresholds (X-axis). Our method has the best performance than the three state-of-the-art methods under the error threshold 50 mm. Our method has the highest percentage of frames that are under error threshold 50 mm (96\%), 3 percentage higher than the second best results with \cite{oberweger15}. Table \ref{tab:compareicvl} shows statistics on several key error thresholds, and our method also achieves significantly better performance than all other methods.

Moreover, We do a qualitative evaluation of our algorithm on NYU and ICVL datasets. Fig. \ref{fig:fig13}(a) shows qualitative results on challenging samples in NYU dataset, and Fig. \ref{fig:fig13}(b) shows qualitative results on random samples in ICVL dataset. Nearly all our estimated poses are kinematically valid, and are close to ground truth, even though there are heavy noise and depth occlusion in the dataset. Our method performs best for most of the testing images in both the datasets. More results are given in the supplementary material.

\begin{table}[h]
\caption{Percentage of frames in the test examples over different error thresholds on NYU hand pose dataset.} \centering
\begin{tabular}{c|c|c|c}
  \hline
  {Threshold(mm)} & $\leq$20 & $\leq$40 & $\leq$50\\
  \hline
  ours  & 18\% & 61\% & 74\%\\
  Zhou et al\cite{zhou2016model} & 9\% & 44\% & 58\%\\
  Tompson et al\cite{tompson14tog} & 1.4\% & 39\% & 53\%\\
  Oberweger et al\cite{oberweger15} & 11\% & 50\% & 65\%\\
  \hline
\end{tabular}
\label{tab:comparenyu}
\end{table}

\begin{table}[h]
\caption{Percentage of frames in the test examples over different error thresholds on ICVL hand pose dataset.} \centering
\begin{tabular}{c|c|c|c}
  \hline
   {Threshold(mm)} & $\leq$20 & $\leq$40 & $\leq$50\\
  \hline
  ours  & 40\% & 91\% & 96\%\\
  Tang et al\cite{tang2014latent} & 18\% & 69\% & 85\%\\
  Zhou et al\cite{zhou2016model} & 47\% & 85\% & 91\%\\
  Oberweger et al\cite{oberweger15} & 45\% & 86\% & 93\%\\
  \hline
\end{tabular}
\label{tab:compareicvl}
\end{table}

\subsection{Ablation Studies}
\begin{figure}[t]
\centering
\includegraphics[width=\linewidth]{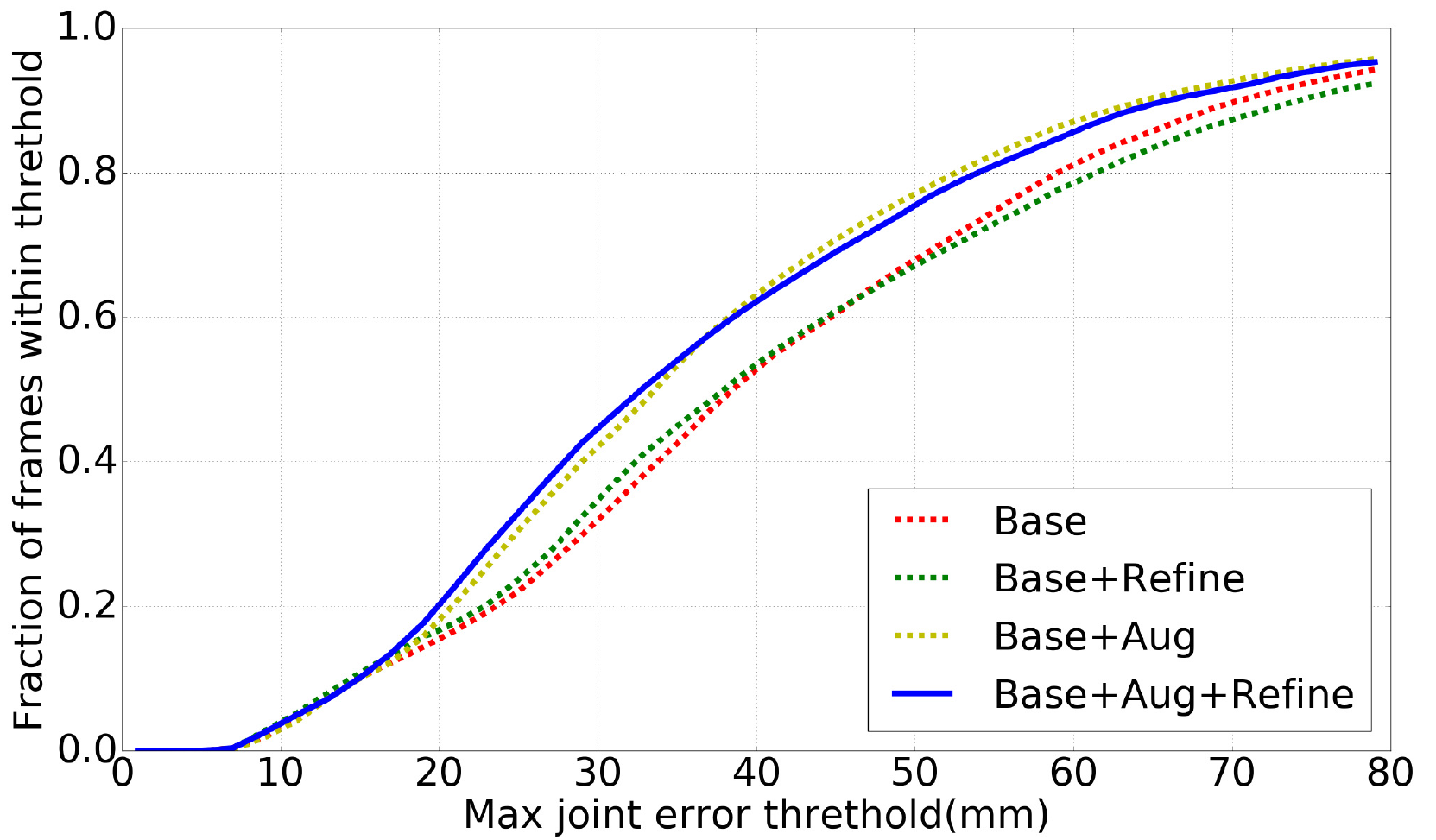}
\caption{Ablation studies with/without data augmentation or TSDF refinement. We show percentage of frames in the test examples under different error thresholds.
Base: pose estimation without data augmentation and TSDF refinement. Base + Refine: pose estimation with TSDF refinement but without data augmentation. Base + Aug: pose estimation with data augmentation but without TSDF refinement. Base + Aug + Refine: pose estimation with data augmentation and TSDF refinement. }
\label{fig:fig9}
\end{figure}

\begin{figure*}[t]
\centering
\begin{tabular}{cc}
\includegraphics[width=\linewidth]{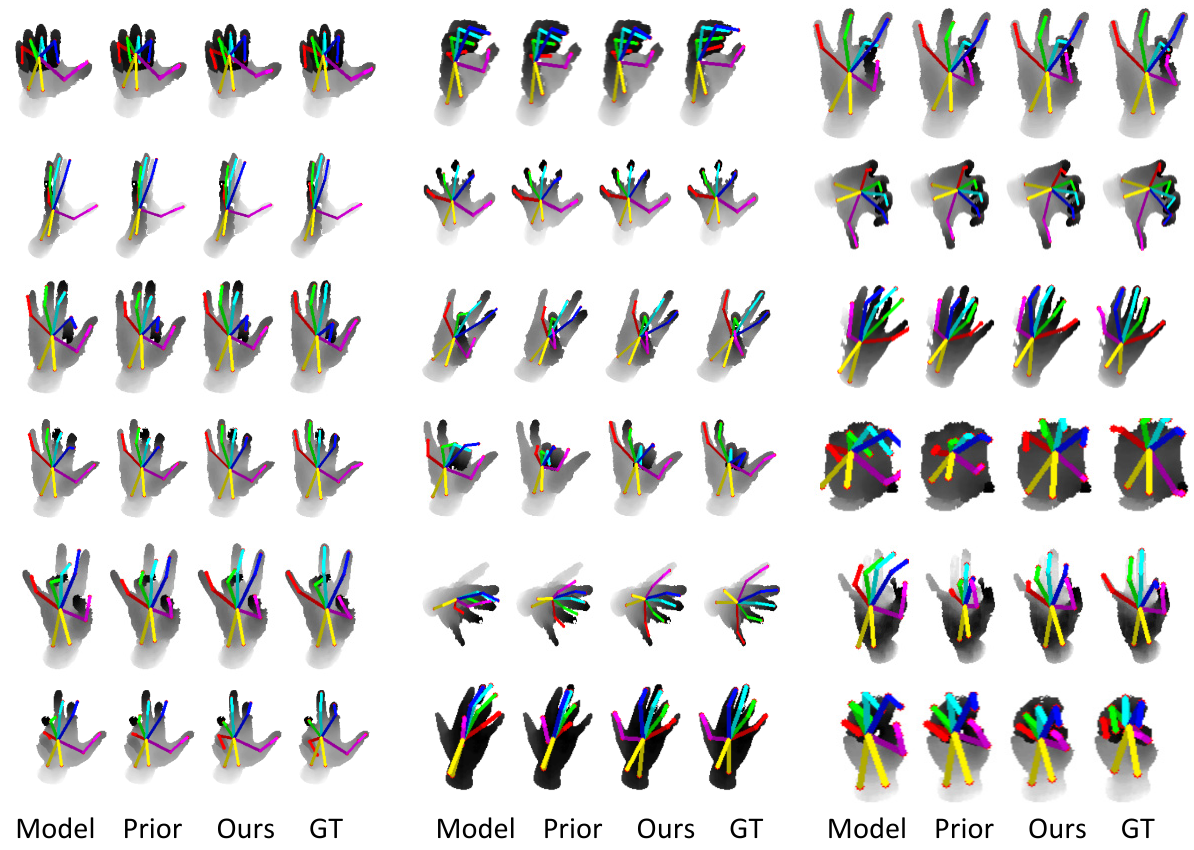} \\
(a) Results on NYU hand dataset \\
\includegraphics[width=\linewidth]{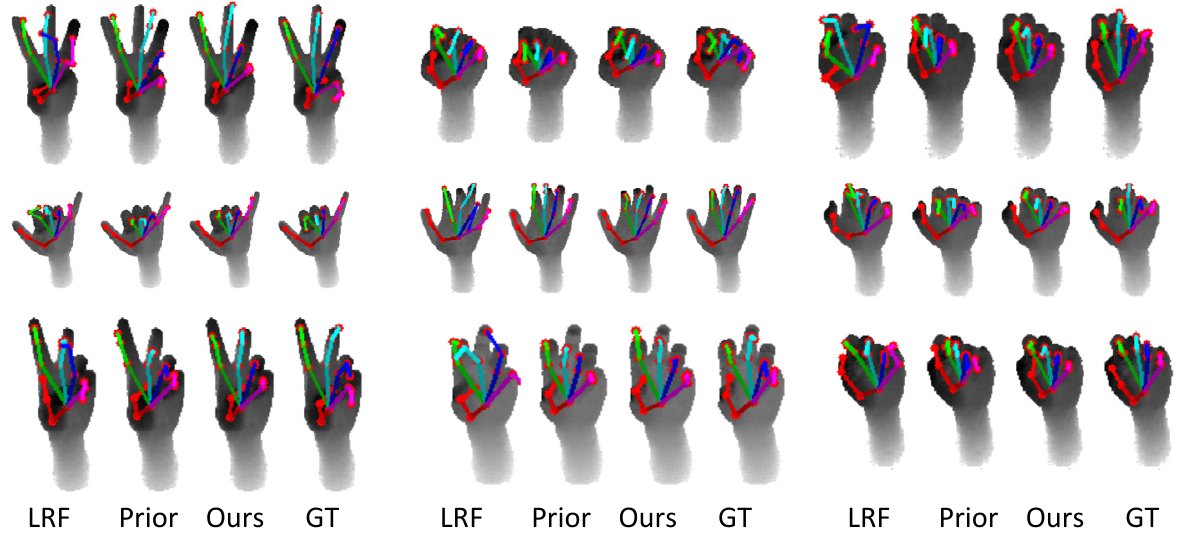}\\
(b) Results on ICVL hand dataset
\end{tabular}
\caption{Qualitative hand pose estimation results on NYU and ICVL hand pose datasets. (a) shows qualitative hand pose results on NYU hand dataset. Comparison with Zhou et al(Model)\cite{zhou2016model}, Oberweger et al(Prior)\cite{oberweger15}, Ours, and Ground truth(GT). (b) shows qualitative hand pose results on ICVL hand dataset.
Comparison with Tang et al(LRF)\cite{tang2014latent}, Oberweger et al(Prior)\cite{oberweger15}, Ours, and Ground truth(GT).}
\label{fig:fig13}
\end{figure*}

To further evaluate our model, we analyze each component of our model and provide quantitative analysis. We evaluate the effect of data augmentation and TSDF refinement for pose estimation on NYU hand pose dataset. Quantitative results are shown in Fig. \ref{fig:fig9}.

\noindent \textbf{Effect of data augmentation.} Hand pose estimation without data augmentation (red and green dashed), as expected, performs inferior as the original NYU hand pose data does not encode hand skeleton size variety. Hand pose estimation with data augmentation significantly improves the performance by 10 percentage under error threshold 50 mm (Base vs. Base + Aug and Base + Refine vs. Base + Aug + Refine). Therefore, data augmentation with different skeleton sizes is critically important for better performance.

\noindent \textbf{Effect of TSDF refinement. }
We also evaluate the impact of TSDF refinement for hand pose estimation.
In Fig. \ref{fig:fig9}, we show the performance of our model with/without TSDF refinement. As we can see, the percentage of test frames with the error within [17.5,37.5]mm improves by about 3\%. Therefore, TSDF refinement helps to especially improves the pose estimation performance under low error threshold, which are more important for accurate pose estimation required by many real applications.

\subsection{Runtime}
Our method is implemented in Python using Keras. We run the experiments on a computer equipped with an Intel Core i7, 32GB of RAM, and a Nvidia GeForce GTX TITAN X GPUs. Our method runs about 30 FPS, which is faster than model-based methods such as 25FPS in \cite{qian2014realtime} and 12FPS in \cite{xu2013efficient}. The runtime of our method is almost the same as \cite{tompson14tog}, while our method is more accurate.

\section{Conclusion}
We present a 3D neural network architecture for hand pose estimation from a single depth image. The input depth image is converted to a 3D volumetric representation, and the 3D pose estimation network is trained directly on it to estimate the location of each hand joint relative to COM of the hand region. The 3D CNN architecture naturally integrates both local 3D feature and global context. Hence, the output of the network is directly in 3D and does not require any further post-processing to integrate context in predefined hand model. To have large training data with better coverage of the hand configuration, we transfer hand pose from existing datasets to handle with variable bone length, and render the depth accordingly. Our method achieves the state-of-the-art performance on both NYU dataset and ICVL dataset.

\section*{Acknowledgment}
The authors appreciate the help and support of Professor Shihong Xia, Professor Kangkang Yin, Professor Jinxiang Chai, Dr. Yichen Wei, Dr. Yangang Wang, Dr. Wenping Zhao, Mr. Markus Oberweger. The authors would like to thank Wenyong Zheng and Jian Shi from ISCAS for help in data augmentation, Kaimo Lin from NUS for helpful collaboration. The authors acknowledge the support of NVIDIA with the donation of the GPUs used for this research.


{\small
\bibliographystyle{ieee}
\bibliography{egbib}
}

\clearpage
\begin{center}
\textbf{\large Supplemental Materials}
\end{center}

\appendix
In this supplementary material, we provide detailed network architecture, explanation of the synthetic data generation, and more results of hand pose estimation.

\section{Network Architecture}
The architecture of our 3D fully convolutional neural network (FCN) for TSDF refinement is shown in Table~\ref{tab:tabletsupp1}(a).
The network consists of an encoder with two layers of 3D convolution + 3D convolution + 3D pooling (with ReLU), and a symmetric decoder with two layers of 3D convolution + 3D convolution + 3D uppooling, followed by three 3D convolution.  For the encoder, we use a 3D convolution layer with kernel size as $3 \times 3 \times 3$, a stride of $(1,1,1)$,  and padding size $(1,1,1)$ to extract features for each level (conv3D layer is followed by ReLU). For the decoder, we up-sample the previous layer¡¯s feature maps, and then concatenate them with the feature map from encoder. We use the same parameters for the convolution layers in decoder. For 3D pooling layer, the kernel size is $2 \times 2 \times 2$ kernel with a stride of $(2,2,2)$. For 3D uppooling layer, we use the same kernel size.

\begin{table*}[t]
\centering

\begin{tabular}{cc}
\begin{tabular}{|c|c|c|c|c|c|}
\hline
layer & channel & kernel size & stride & padding & Activation \\ \hline
conv3D & 32  & (3,3,3) &(1,1,1)&(1,1,1)  &ReLU  \\ \hline
conv3D & 32  & (3,3,3) &(1,1,1)&(1,1,1)  &ReLU  \\ \hline
pooling3D & -  & (2,2,2) &(2,2,2)  & - & - \\ \hline
conv3D & 64  & (3,3,3) &(1,1,1)&(1,1,1)  &ReLU  \\ \hline
conv3D & 64  & (3,3,3) &(1,1,1)&(1,1,1)  &ReLU  \\ \hline
pooling3D & -  & (2,2,2) &(2,2,2)  & - & - \\ \hline
conv3D & 128  & (3,3,3) &(1,1,1)&(1,1,1)  &ReLU  \\ \hline
conv3D & 128  & (3,3,3) &(1,1,1)&(1,1,1)  &ReLU  \\ \hline
upPooling3D & -  & (2,2,2) &-  & - & - \\ \hline
conv3D & 64  & (3,3,3) &(1,1,1)&(1,1,1)  &ReLU  \\ \hline
conv3D & 64  & (3,3,3) &(1,1,1)&(1,1,1)  &ReLU  \\ \hline
upPooling3D & -  & (2,2,2) &-  & - & - \\ \hline
conv3D & 32  & (3,3,3) &(1,1,1)&(1,1,1)  &ReLU  \\ \hline
conv3D & 16  & (3,3,3) &(1,1,1)&(1,1,1)  &ReLU  \\ \hline
conv3D & 1  & (3,3,3) &(1,1,1)&(1,1,1)  & Tanh  \\ \hline
\end{tabular} \\
(a) Network architecture for TSDF refinement. \\ \\
\begin{tabular}{|c|c|c|c|c|c|}
\hline
layer & channel & kernel size & stride & padding & Activation \\ \hline
conv3D & 8  & (5,5,5) &(1,1,1)&(2,2,2)  &ReLU  \\ \hline
pooling3D & -  & (2,2,2) &(2,2,2)  & - & - \\ \hline
conv3D & 8  & (5,5,5) &(1,1,1)&(2,2,2)  &ReLU  \\ \hline
pooling3D & -  & (2,2,2) &(2,2,2)  & - & - \\ \hline
FC & 2048 & - &-  & - & ReLU \\ \hline
dropout & \multicolumn{5}{c|}{dropout fraction=0.5} \\ \hline
FC & 2048 & - &-  & - & ReLU\\ \hline
dropout & \multicolumn{5}{c|}{dropout fraction=0.5} \\ \hline
FC & 42 & - &-  & - & - \\ \hline
\end{tabular}\\
(b) Network architecture for hand pose estimation.
\end{tabular}

\caption{Network architectures for TSDF refinement and pose estimation. (a) shows the network architectures for TSDF refinement. (b) shows network architecture for hand pose estimation. The channel number of the last FC is equal to $3\times$\# of joints. This table shows the case for NYU hand pose dataset, where \# of joints = 14. For ICVL dataset, the channel number of the last FC is 48. }
\label{tab:tabletsupp1}

\end{table*}


Our 3D ConvNet architecture for pose estimation is shown in Table~\ref{tab:tabletsupp1}(b). The network starts with two layers of 3D convolution + 3D pooling + ReLU, followed by three fully connected layers. The last fully connected layer produces a feature of dimension $3\times$ \# of joints, which is used to directly estimate the 3D location of each joint relative to the COM. In order to increase the receptive field with a relatively shallower architecture, we set a larger kernel size, $5 \times 5 \times 5$, with a padding of $(2,2,2)$.
We add dropout layers to prevent overfitting.



\section{Synthetic Data Generation}
In this section, we introduce more details about the synthetic data generation.

\subsection{Pose Parametrization}
Not as in most of the dataset where a hand pose is represented by the specific locations of each joint, we use the location of palm, the length of each bone, and the angle on each joint to represent a pose \cite{parent2012computer}.
This representation encodes the bone length invariance for multiple poses from a single subject, and allows the joint angle to be easily transfered to other hand models.

\begin{figure}[t]
\centering
\begin{tabular}{cc}
\includegraphics[width=0.3\linewidth]{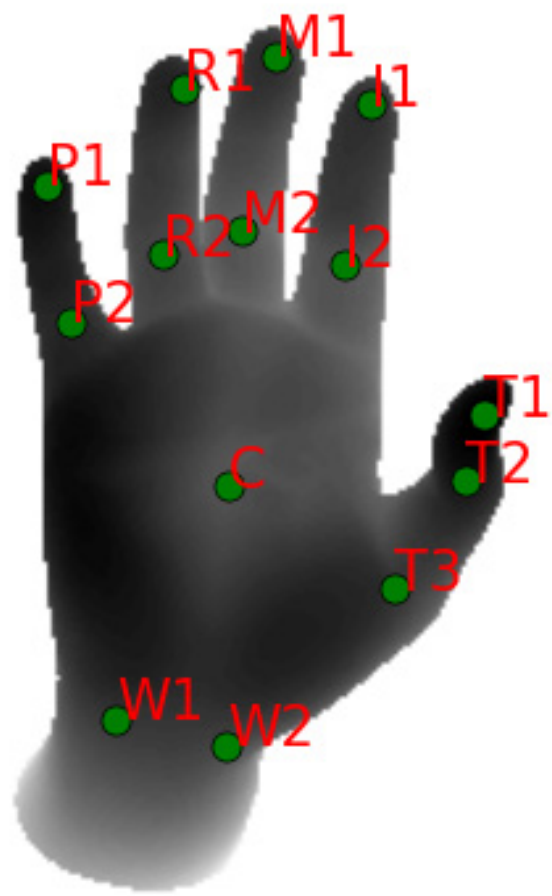} & \includegraphics[width=0.3\linewidth]{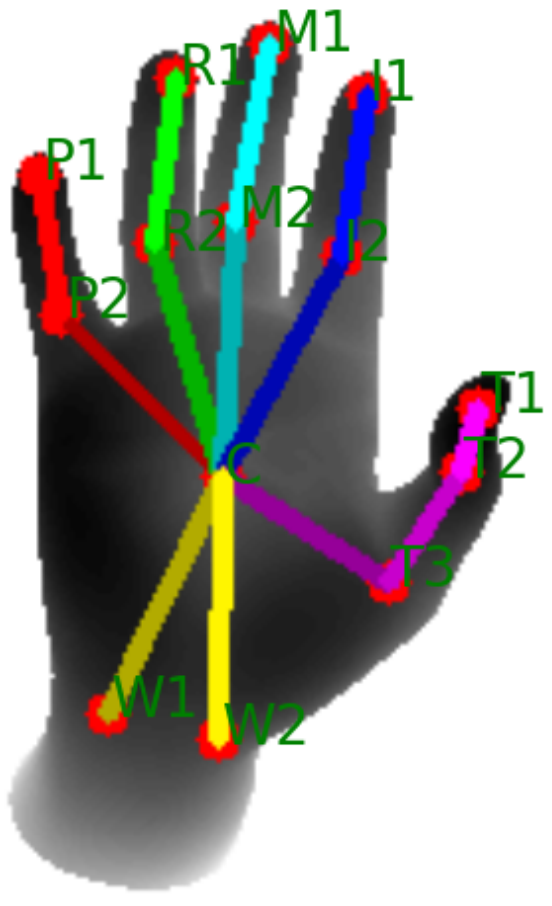}\\
(a) Hand skeleton joints & (b) Joint links
\end{tabular}
\caption{Illustration of hand skeleton model and joint links (NYU hand pose dataset). (a) shows the hand skeleton joints. (b) shows the joint links. }
\label{fig:suppfig1}
\end{figure}

The training set of the NYU hand dataset are collected by one subject, which is perfect for learning joint angle since the bone length are fixed.
However, the annotation is not accurate enough to satisfy the bone length invariance. As such, we learn the bone length and use a forward kinematic model to recover the joint angles from 3D joint locations. Hereafter, we use the following denotations: C: palm center, T: thumb, I: index, M: middle, R:ring, P:pinky, W1: wristside1, W2: wristside2. Fig. \ref{fig:suppfig1} illustrates the hand joints.

\begin{figure}[t]
\centering
\includegraphics[width=\linewidth]{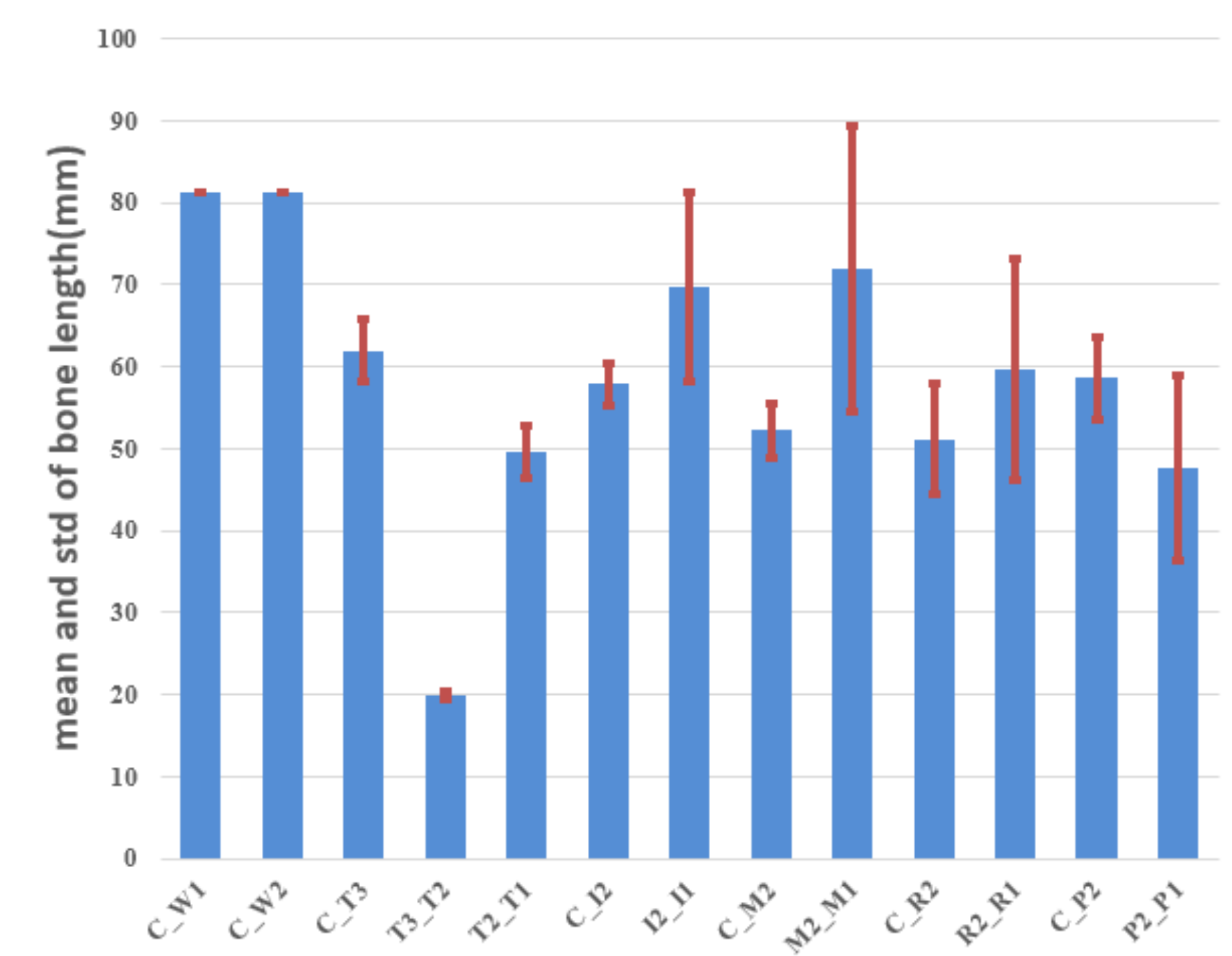}
\caption{Mean and standard deviation of selected bone lengths (NYU hand pose dataset). The standard deviation is relatively large for joints on fingers. As such, the joint angles directly calculated via the joint locations may not be accurate.}
\label{fig:suppfig2}
\end{figure}

We first learn the bone length of the subject in the training set of the NYU hand dataset. For each training data, we calculate the length of each bone as the distance between the two joints on its end. The typical distribution of some bones are shown in Fig. \ref{fig:suppfig2}. We choose the mean of each bone length distribution as the final length, and use it in the following optimization to recover joint angles.

On each training image, the goal of the optimization is to estimate a set of joint angles, such that the joint position calculated together with the bone length is close to the ground truth annotation:
\begin{equation}
\mathbf{p} = argmin_p \sum_i ||R_{i}(\mathbf{p})-O_{i}||^2
\end{equation}
\noindent where $\mathbf{p}$ stands for a hand pose consisting of the hand root joint position and relative angles of individual joints. $\{O_i\}_{i=1}^n$ are the coordinates of ground truth hand joint locations. $\{R_i=f_{i}(\mathbf{p})\}_{i=1}^n$ are the coordinates of hypothesized hand joint locations, which are computed by hand skeleton model and forward kinematics (refer to Eqn.~\ref{eqfk} for $f_{i}(\mathbf{p})$), $n$ is the number of hand joints.

For NYU hand dataset \cite{tompson14tog}, the root position is set to the palm center joint ${O(C)}$, and the root orientation $\mathbf{R} = [\mathbf{r}_x,\mathbf{r}_y,\mathbf{r}_z]$ is computed as follows:
\begin{eqnarray}
\nonumber \mathbf{r}_x &=& {(O(W1)-O(C)) \times (O(W2)-O(C))}\\
\nonumber \mathbf{r}_z &=& {O(M1)-O(C)} \\
\nonumber \mathbf{r}_y &=& {\mathbf{r}_z \times \mathbf{r}_x}
\end{eqnarray}
\noindent where $O(.)$ is the position of a joint. $M1$: the root joint of middle finger.

The relative offset to its ancestor joint of a joint that links to the root joints can be computed by ${(O(.) - O(C))}\mathbf{R}$. For a joint $k$ on a finger, its relative offset to its ancestor can be computed as $[0,0,||O(k) - O(ancestor(k))||_2]$.

The hand kinematic chain is modeled as a set of $n$ joints linked by segments whose connectivity are expressed in terms of a local coordinate system $\theta_i(i=1,...,n)$. The coordinate transformation from $\theta_i$ to
its ancestor joint $\theta_{i-1}$ is described by a $4\times4$ rotation-translation matrix $\mathbf{D}_{\theta_{i-1}}^{\theta_{i}}$\cite{Cerveri2003377}, which can be calculated with joint angles and the relative offset. Therefore, the coordinates of a joint $k$ in the world coordinate system can be recovered computed as follows
\begin{equation}
\label{eqfk}
R_k(\mathbf{p}) = f_k(\mathbf{p}) = \mathbf{D}_{root}^{\theta_{1}}\mathbf{D}_{\theta_{1}}^{\theta_{2}}...\mathbf{D}_{\theta_{k-1}}^{\theta_{k}}
\end{equation}
\noindent where $f_k(\mathbf{p})$ is the forward kinematic function for joint $k$, $\mathbf{p} = \{\theta_1,...,\theta_n\}$ is the parametrized hand pose.

In order to recover the joint angle with 3D joints, we use inverse kinematic (IK) optimization, which optimizes the parametrized hand pose $\mathbf{p}$ for the smallest forward kinematics cost, that is, the pairwise distances between the hand joint positions in hand model (computed with (\ref{eqfk})) and the corresponding ground truth hand joints. During IK, the position and orientation of root joint are fixed. For other joints, we enforce the joint angles to be within their valid ranges (similar to \cite{SimoSerraPFC2011}) during IK. The IK optimization is solved with particle swarm optimization (PSO) \cite{pso} for implementation convenience, yet other nonlinear optimization algorithms \cite{wang2009real} can also be used.

\subsection{Pose Transfer}
We encode the recovered hand pose into a standard Biovision hierarchical data (BVH) file \cite{bvh}, which consists of the position of root joint, relative angles of each joint, and canonical bone lengths. We modify the relative offsets in the BVH file to generate models with different bone lengths, use linear skinning techniques \cite{baran2007automatic} to animate the chosen 3D hand CAD model given the pose, and then render depth images with commercial software Autodesk Maya.

\section{More Results}
\begin{figure*}[ht]
\centering
\begin{tabular}{cc}
\includegraphics[width=\linewidth]{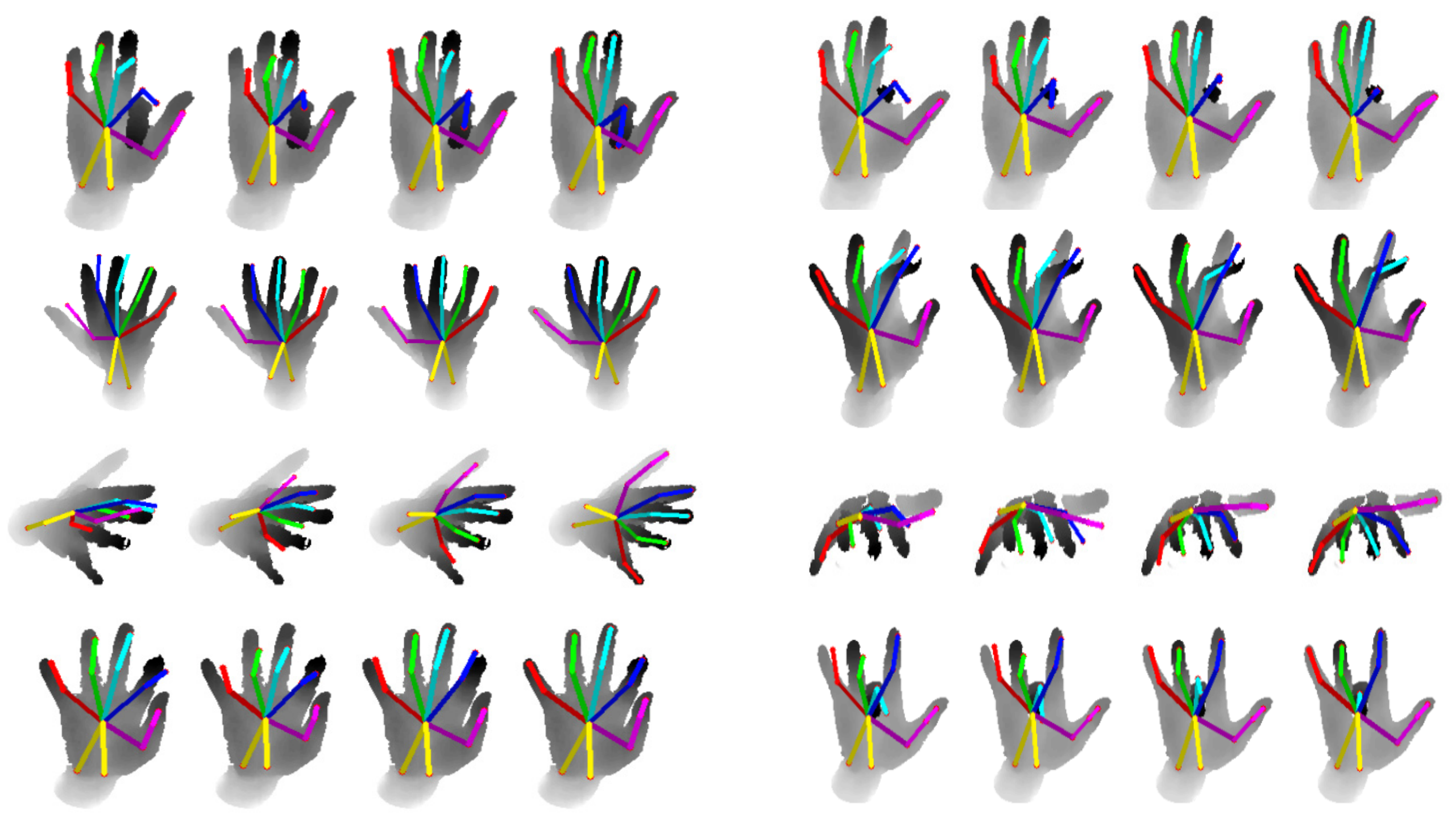}\\
\includegraphics[width=\linewidth]{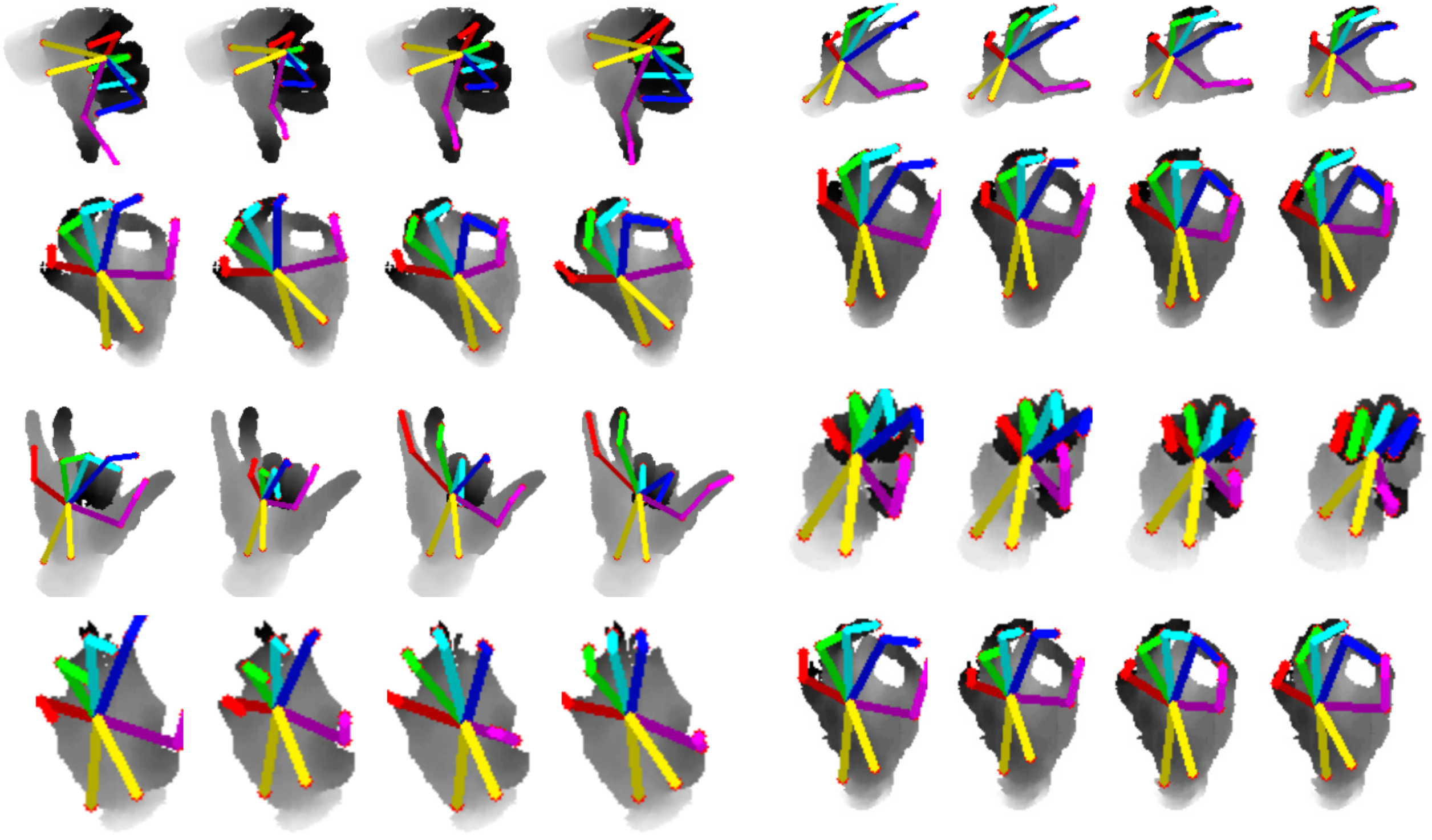}\\
\includegraphics[width=\linewidth]{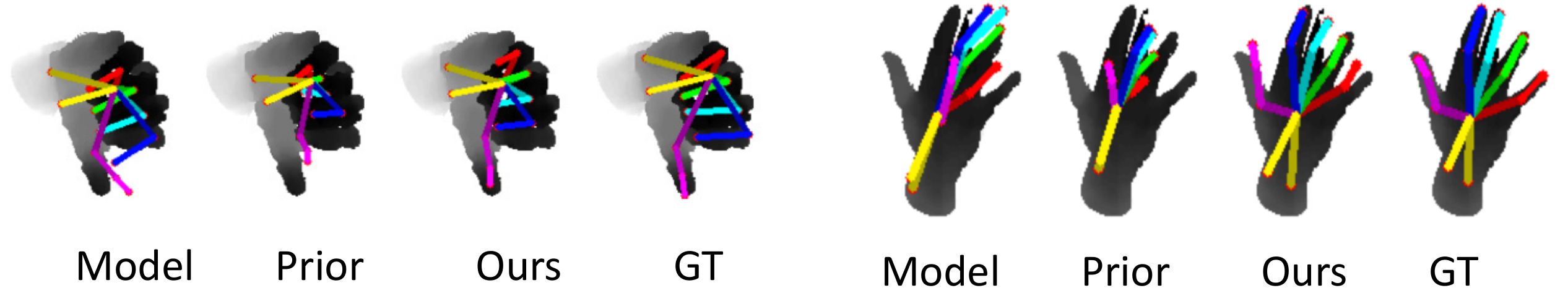}\\
\end{tabular}
\caption{More results on NYU hand pose dataset. shows qualitative hand pose results on NYU hand dataset. Comparison with Zhou et al(Model)\cite{zhou2016model}, Oberweger et al(Prior)\cite{oberweger15}, Ours, and Ground truth(GT).}
\label{fig:suppfig4}
\end{figure*}

\begin{figure*}[t]
\centering
\includegraphics[width=\textwidth]{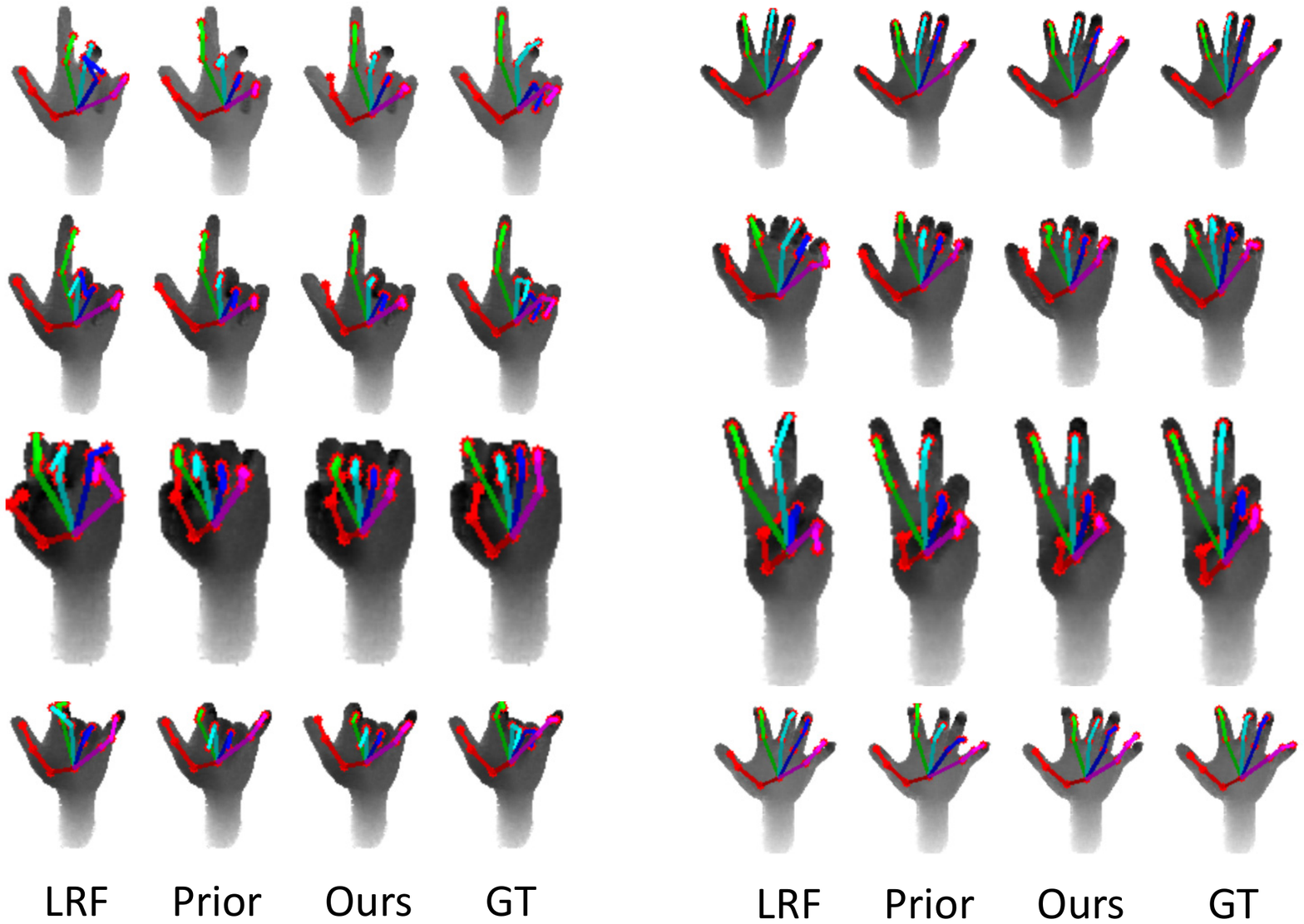}
\caption{More results on ICVL hand pose dataset. shows qualitative hand pose results on ICVL hand dataset.
Comparison with Tang et al(LRF)\cite{tang2014latent}, Oberweger et al(Prior)\cite{oberweger15}, Ours, and Ground truth(GT).}
\label{fig:suppfig5}
\end{figure*}

\begin{figure*}[b]
\centering
\includegraphics[width=\textwidth]{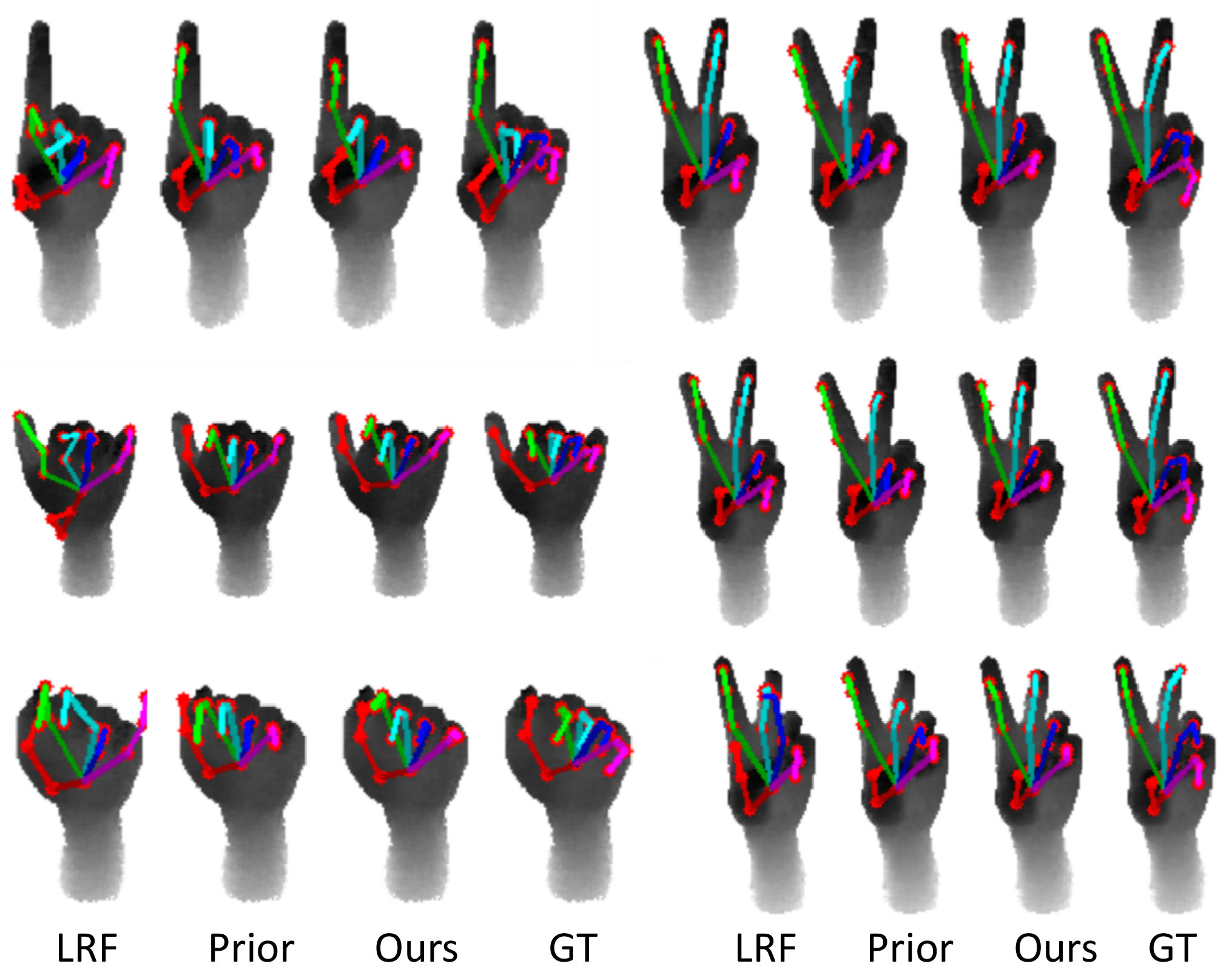}
\caption{Examples of wrongly annotated data in ICVL pose dataset. The ground truth annotations provided by \cite{tang2014latent} are often not kinematically valid for Index, Middle, Ring and Pinky if they are self-occluded. Above results demonstrates that our model can always recover valid poses. }
\label{fig:suppfig6}
\end{figure*}

We show more visual comparison on images from NYU hand pose dataset and ICVL hand pose dataset (refer to Supplementary Video for detail). Fig. \ref{fig:suppfig4} shows qualitative hand pose results on NYU hand dataset. Comparison with Zhou et al(Model)\cite{zhou2016model}, Oberweger et al(Prior)\cite{oberweger15}, Ours, and Ground truth(GT). Fig. \ref{fig:suppfig5} shows qualitative hand pose results on ICVL hand dataset. Comparison with Tang et al(LRF)\cite{tang2014latent}, Oberweger et al(Prior)\cite{oberweger15}, Ours, and Ground truth(GT). The joint annotation in ICVL hand dataset is inaccurate, while our method can learn a robust model, which gives accurate results for many frames even though the ground truth annotations are inacurate. Fig. \ref{fig:suppfig6} shows such results.

\end{document}